\documentclass[11pt]{article}
\usepackage{coling2020}
\usepackage{times}
\usepackage{url}
\usepackage{latexsym}
\usepackage{times}
\usepackage{latexsym}
\usepackage{mathtools}
\usepackage{bbm}
\usepackage{dutchcal}
\usepackage{graphicx}
\usepackage{amsmath,bm}
\usepackage{multirow}
\usepackage{subcaption}

\usepackage{microtype}
\usepackage{tabularx}
\usepackage[ruled, lined, linesnumbered, commentsnumbered, longend]{algorithm2e}
\usepackage{algorithmic}
\usepackage[bottom]{footmisc}

\newcommand{\model}[1]{\textsc{#1}\xspace}

\usepackage{xcolor}

\colingfinalcopy 

\title{\model{CosMo}:~Conditional \textsc{Seq2Seq}-based Mixture Model for Zero-Shot Commonsense Question Answering}

\author{Farhad Moghimifar$^1$ \and \bf{Lizhen Qu}$^2$ \and \bf{Yue Zhuo}$^3$ \\
        \bf{Mahsa Baktashmotlagh}$^1$ \and \bf{Gholamreza Haffari}$^2$ \\
         $^1$The School of ITEE, The University of Queensland, Australia\\
         $^2$Faculty of Information Technology, Monash University, Australia\\
         $^3$School of CSE, The University of New South Wales,  Australia\\
         \tt \{f.moghimifar,m.baktashmotlagh\}@uq.edu.au\\
         \tt firstname.lastname@monash.edu, 
         \tt terry.zhuo@unsw.edu.au
         }

\date{}

\begin{document}
\maketitle

\begin{abstract}
Commonsense reasoning refers to the ability of evaluating a social situation and acting accordingly. Identification of the implicit causes and effects of a social context is the driving capability which can enable machines to perform commonsense reasoning. The dynamic world of social interactions requires context-dependent on-demand systems to infer such underlying information. However, current approaches in this realm lack the ability to perform commonsense reasoning upon facing an unseen situation, mostly due to incapability of identifying a diverse range of implicit social relations. Hence they fail to estimate the correct reasoning path.
In this paper, we present Conditional \textsc{Seq2Seq}-based Mixture model~(\model{CosMo}), which provides us with the capabilities of dynamic and diverse content generation. We use \model{CosMo} to generate context-dependent clauses, which form a dynamic Knowledge Graph~(KG) on-the-fly for commonsense reasoning.
To show the adaptability of our model to context-dependant knowledge generation, we address the task of zero-shot commonsense question answering. The empirical results indicate an improvement of up to +5.2\% over the state-of-the-art models.    
\end{abstract}

\section{Introduction}
\label{sec:intro}
\begin{figure}[h]
    \centering
    \includegraphics[width=\linewidth]{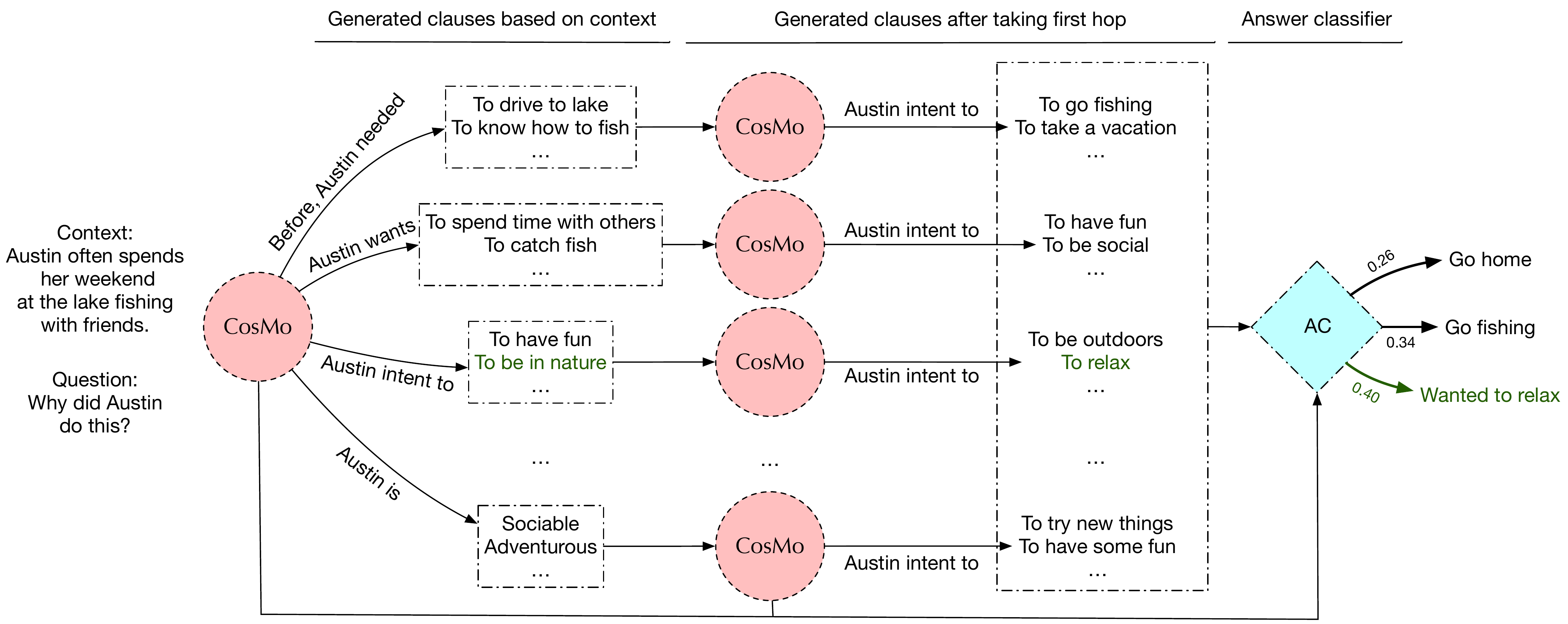}
    \caption{The illustration of the process of answering a question with our proposed model. The model receives a context and a question. Using Conditional \textsc{Seq2Seq}-based Mixture Model~(\model{CosMo}), clauses based on the context and different relations are generated. The generated clauses are then used to generate new set of clauses based on the question. The Answer Classifier~(AC) module selects the answer with higher score based on generated clauses at the last step and scores of \model{CosMo}. The path to the correct answer is distinguished with green color. }
    \label{fig:model}
\end{figure}

People understand narratives of everyday life by capitalising
on their commonsense knowledge. They can easily reason about unobserved causes and effects in relation to the events described in narratives, as well as plausible characteristics and mental states of the involved persons. Although this kind of reasoning seems trivial for humans, it is still out of reach for current natural language understanding~(NLU) systems. Recently, there have been fast-growing interests in building AI systems with such human-like reasoning capabilities based on inferential commonsense knowledge~\cite{storks2019commonsenseSurvey,bosselut2019zeroShotQA}. Such systems are often evaluated by answering questions based on narratives. As illustrated in Figure \ref{fig:model}, given a narrative ``Austin often spends her weekend at the lake fishing with friends" and a question regarding the intention of Austin, an AI system is supposed to associate this event to relevant events in an inferential knowledge base or a web-scale corpus, find plausible reasons of those events, and conclude that ``wanted to relax" is the most probable answer. 

Knowledge-based approaches to such commonsense question-answering (QA) require an inferential knowledge base. ATOMIC~\cite{sap2019atomic} is the largest commonsense knowledge base of this kind, which contains 300,000 short textual description of events and 877,000 typed \textit{if-then} relations between events, categorised into 9 dimensions. For instance, \textit{IF} the event ``X puts trust in Y" occurs and the target relation is ``xWant", \textit{THEN} ``X wants to develop a relationship". Prior work utilizes knowledge in ATOMIC by formulating the learning problem as event prediction in \textit{if-then} relations~\cite{bosselut2019comet}. In particular, they encode the textual description of an event and a relation into an embedding, and maximise the probability of predicting the description of the associated event or characteristic of an involved person. However, due to the nature of commonsense knowledge, given an event and a relation, there are multiple plausible associated events. Moreover, these models fail to predict all associated events to a given event and relation, hence fail to identify every implicit reasoning path to address the task of question answering.

To address the above challenges, we propose a conditional \textsc{Seq2Seq}-based mixture model, named \model{CosMo}, to answer questions on everyday inferential knowledge in a \textit{zero-shot} setting. As the events in our target commonsense KGs are described in text, such as ATOMIC, we distil the knowledge into a Sequence-to-Sequence~(\textsc{Seq2Seq}) model. The distilled model memorises knowledge of the KG and generalise it to new events for a given context and specific relation, on-demand. However, a direct application of maximum-likelihood training on \textsc{Seq2Seq} models leads to deteriorated performance, because the underlying distribution over \textit{diverse} outputs is inherently \textit{multi-modal}~\cite{shen2019mixture}. To address the above challenge, we incorporate a latent variable into a pre-trained \textsc{Seq2Seq} transformers~\cite{yan2020prophetnet}. Each value of the variable corresponds to an embedding, which indicates the hidden factors explaining different hypothesis. Unlike the existing training methods of such models that cannot guarantee a definitive alignment of different hypothesis with different components during training, our proposed constrained-EM~(expectation-maximisation) training procedure enforces that for the same input, different model outputs align with different latent embedding. 

To tackle the task of Commonsense QA, we use \model{CosMo} to generate context-dependent information, for desired relations. The new generated events form a dynamic knowledge graph, which is reasoned over, until the correct answer is chosen by the model. The ubiquitous nature of everyday commonsense knowledge and lack of training data for each situation motivate addressing the task in zero-shot setting. It is unrealistic to expect the presence of manually constructed training datasets in each commonsense QA domain. With the lack of training data in such question-answering settings, we devise a bespoken answer scoring module to assess the likelihood of each answer. Given a narrative and a question, the model generates sequences of plausible fact descriptions, as \emph{reasoning paths}, by choosing different values of the latent variable. For each answer and each reasoning path, the scoring module computes a score based on similarity between the answer and the last fact description in the reasoning path. The most probable answer is determined by its answer score and the probability of the associated reasoning path \textit{jointly}.

To sum up, our contributions are three-folds:
\begin{itemize}
    \item We propose \model{CosMo}, a conditional \textsc{Seq2Seq}-based mixture model for zero-shot question answering on inferential knowledge. \model{CosMo}  constructs a dynamic KG on-demand, which is used to reason over and answer commonsense questions.
    \item We propose a novel training procedure to enforce hard alignments between latent variables and diverse hypotheses, to ensure the diversity of the associated KG events  distilled in the \textsc{Seq2Seq}. 
    \item Our experimental results on SocialIQA~\cite{sap2019socialiqa} show that our model achieves superior accuracy and significantly more diverse hypotheses than competitive baselines. 
\end{itemize}

\section{Related Works}
\label{sec:related_works}
\paragraph{Commonsense Knowledge Bases}
Introducing Commonsense Knowledge Graphs (KG) has provided a source of information for machines on the task of commonsense reasoning. This KGs, encode commonsense relations between different pair of events/concepts.
Speer et al.~\shortcite{speer2017conceptnet} assembled a KG from a variety of sources, ConceptNet, which represents general knowledge in form of tuples. While ConeptNet is more centered around taxonomic knowledge, Sap et al.~\shortcite{sap2019atomic} constructed ATOMIC, a KG consisting inferential knowledge. Information in ATOMIC~\cite{sap2019atomic} are presented as \emph{if-then} relationship between events, mental states, and persona. The information in ATOMIC includes such information for the agent of the event and others, who might be affected. Zhang et al.~\shortcite{zhang2020aser} proposed an automatic approach for collecting eventuality KG, ASER, consisting events, activities, and states.
Although the presented KGs provide rich commonsense information, reasoning about social situation requires a dynamic on-demand approach for context-dependent information generation.

\paragraph{Commonsense Knowledge Base Completion}
As an essential way of enabling machines to perform commonsense reasoning, the methods for automatic KG construction and completion have been studied recently. Sap et al.~\shortcite{sap2019atomic} used LSTM as a generator for commonsense knowledge about social situations. Davison et al.~\shortcite{davis2015commonsense} proposed an unsupervised method to extract commonsense knowledge from pre-trained models, to complete KG. Malaviya et al.~\shortcite{malaviya2020commonsense} developed a model which takes both structural and semantic characteristics of the nodes in a KG to address this task. On the other hand, some works have developed generative models on top of pre-trained language models to extract new commonsense information~\cite{bosselut2019comet,malaviya2019exploiting}. However, when adapting to the KG, the previously acquired knowledge of these models is forgotten. Unlike these approaches,  our neural model ensures both diversity and memorisation.

\paragraph{Commonsense Question Answering} Recent surge of commonsense question answering dataset~\cite{sap2019socialiqa,zellers2018swag} has led to many supervised approaches to address this task. Most of these approaches are based on transfer learning, where a large scale pre-trained model~\cite{lan2019albert,devlin2018bert} is finetuned on the target task\cite{sap2019socialiqa}. On the other hand, with the release of new commonsense KGs, such as ATOMIC~\cite{sap2019atomic} and ConceptNet~\cite{speer2017conceptnet}, the possibility of enriching language models with these KGs has been investigated vastly~\cite{lv2020graph,mitra2019exploring,banerjee2020knowledge}. Most of these approaches map the context of a question to an entity/event in KG and perform reasoning on the KG~\cite{weissenborn2017dynamic,paul2019ranking,lin2019kagnet}. While these methods enable the system to perform multi-hop reasoning, they are limited to the set of entities in KG. 
Other works deployed generative models to generate context-aware information to answer the questions~\cite{shwartz2020unsupervised,bosselut2019zeroShotQA,banerjee2020knowledge}. These approaches overcome the limitation of static KGs, but they lack diversity in generating new relations, which makes their inference path limited. Furthermore, for scoring the answer choices they have solely relied on the conditional likelihood of the generative model, which lack to perform when the distribution of KG and QA differ.

\paragraph{Zero-shot Question Answering}
In recent years, zero-shot learning has become a popular method to conquer the inability of machine learning system to perform on unobserved data~\cite{wang2019survey}. While this method has been thoroughly researched in other fields~\cite{zhao2019towards}, the necessity of using such approach has inspired many works in question answering systems as well. Visual Question Answering
~\cite{teney2016zeroshotVQA} adopted a zero-shot learning approach to extract features from unseen text description about given images. Lewis~\shortcite{lewis2019unsupervisedQA} proposed an unsupervised extractive question answering model by using unsupervised data generation for converting the QA task to a cloze translation task. Puri~\shortcite{puri2020trainingQAsyntheticData} improved the quality of unsupervised extractive question answering systems by introducing an approach involving answer generation, question generation and roundtrip filtration. In addition, Li et al.~\shortcite{li2020harvestingQAPairsForUnsupervisedQA} used Wikipedia's data in order to overcome the drawback of Lewis et al.~\shortcite{lewis2019unsupervisedQA}. 
Compared to the previous question answering model, our model shows the novelty in generation new clauses from given contexts.

\section{Our Approach}
\label{sec:models}
In this section, we present our model~\model{CosMo} for question answering in the zero-shot setting. In this setting, there is no training data for the QA task, thus we train our model only on ATOMIC to acquire inferential knowledge. We apply the trained model to answer multi-choice questions on everyday inferential knowledge by augmenting the trained model with a non-parametric answer scoring module.

Formally, given a narrative describing a context $\bm{c}$, a commonsense-required question $\bm{q}$, and a set of $m$ candidate answers $\mathcal{A} = \{\bm{a}^0, ..., \bm{a}^m\}$, the task is to choose the most plausible answer from the set $\mathcal{A}$. In order to measure the plausibility of answer candidates, the required commonsense inferential knowledge is from a large external knowledge base $\mathcal{B}$, which is a set of typed \textit{if-then} relations. Each \textit{if-then} relation takes the form of ``\textit{if} $\bm{z}_i$ \textit{and} $r$ \textit{then} $\bm{z}_j$", where $\bm{z}_i$ denotes a word sequence describing that the event $i$ and $r$ is a relation between events from the pre-defined set $\mathcal{R}$. Such a relation is also referred to as inference dimension in ATOMIC.

To this end, we define the target task as finding the most plausible answer by applying inferential reasoning over the knowledge base $\mathcal{B}$. The reasoning process is characterised by finding a plausible reasoning path $\bm{Z}$, which starts from a given context $\bm{c}$, along the target relation determined by a question $\bm{q}$, to reach an answer $\bm{a}$. A reasoning path is a sequence of events correlated by \textit{if-then} relations derived from $\mathcal{B}$. Therefore, for a given context $\bm{c}$ and a question $\bm{q}$, we find the most likely answer by solving the following optimisation problem:

\begin{equation}
    \arg\max_{\bm{a} \in \mathcal{A}, \bm{Z} \in \mathcal{Z}} \log Pr(\bm{a}| \bm{Z}) Pr(\bm{Z} | \bm{c}, r, \mathcal{B}) Pr(r | \bm{q})
    \label{eq:model_score}
\end{equation}
where $Pr(r | \bm{q})$ is the probability of a target relation $r$ given the question~\footnote{Due to the regularity of questions in the benchmark dataset, we can directly apply rules to find out the target relations. Thus $Pr(r | \bm{q})$ is always one in our experiments.}; $Pr(\bm{a}| \bm{Z})$ denotes an \textit{answer scoring module} estimating the probability of an answer $\bm{a}$ given a reasoning path; $Pr(\bm{Z} | \bm{c}, r, \mathcal{B})$ is the probability of a reasoning path $\bm{Z}$, conditioned on the context $\bm{c}$, the target relation $r$, and the knowledge base $\mathcal{B}$. The local distribution $Pr(\bm{Z} | \bm{c}, r, \mathcal{B})$ can be further factorized into
\begin{align}
    Pr(\bm{Z} | \bm{c}, r, \mathcal{B}) &= Pr(\bm{z}_{0}|\bm{c}, r, \mathcal{B}) \prod_{t=1}^T Pr(\bm{z}_{t}| \bm{z}_{<t}, \bm{c}, r, \mathcal{B})\\
                    &= Pr(\bm{z}_{0}|\bm{c}, r, \mathcal{B}) \prod_{t=1}^T Pr(\bm{z}_{t}| \bm{z}_{t-1}, r, \mathcal{B})
\end{align}
where $Pr(\bm{z}_{0}|\bm{c}, r, \mathcal{B})$ is the distribution of  the first event given a context and a target relation and $Pr(\bm{z}_{t}| \bm{z}_{<t}, \bm{c}, r, \mathcal{B})$ characterises the distribution of  future events. To simplify inferential reasoning, we assume $\bm{z}_{t}$ with $t > 0$ is conditionally independent of $\bm{c}$ and $\bm{z}_{< t-1}$ such that both $Pr(\bm{z}_{t}| \bm{z}_{t-1}, r, B)$ and $Pr(\bm{z}_{0}|\bm{c}, r, \mathcal{B})$ can be estimated by the same module, which predicts a future event by taking a textual description and the target relation as input. The module is referred to as the \textit{KB module} because it is trained on ATOMIC to acquire inferential knowledge. 

In the following, we will detail the KB module as well as its training procedure on ATOMIC, followed by presenting the answer scoring module and how to apply them together for the target task.

\subsection{KB Module for Inferential Reasoning}
\label{subsec:KGN}
The KB module aims to encode \textit{if-then} relations in the KB $\mathcal{B}$ into model parameters, and apply the knowledge to infer future events given a target relation and a text describing a current event or a context. As both inputs and outputs are word sequences, we formulate the task of event prediction as a sequence to sequence prediction problem. As a result, we are able to exploit the powerful pre-trained transformer based \textsc{Seq2Seq} models~\cite{devlin2018bert,yan2020prophetnet} as the backbone.

The key challenge of using pre-trained \textsc{Seq2Seq} models on ATOMIC is the diversity of output events based on a given event $z_i$ and relation $r$. The key idea herein is to align different latent factors $h_k$ with different outputs for the same input. Each latent factor is the value of a latent variable for alignment. After introducing the latent variable, we obtain a conditional mixture model of the following form:
\begin{align}
    Pr(\bm{z}|\bm{x}, r;\bm{\theta}_{\mathcal{B}}) = \sum_{k=1}^K Pr(\bm{z}, h_k |\bm{x}, r; \bm{\theta}_{\mathcal{B}}) = \sum_{k=1}^K Pr(\bm{z}| h_k, \bm{x}, r;\bm{\theta}_{\mathcal{B}})Pr(h_k|\bm{x})
\end{align}
where $\bm{x}$ denotes either the description of an event or a context. Here we replace $\mathcal{B}$ with the model parameters $\bm{\theta}_{\mathcal{B}}$ as the knowledge-base is encoded into the model parameters. We further assume $Pr(h_k|\bm{x})$ follows a uniform distribution during prediction because target QA datasets follow a more different distribution than ATOMIC.  

As each latent variable value can be represented by a symbol,  the module for $Pr(\bm{z}| \bm{h}_k, \bm{x}, r)$ is realized by a \textsc{Seq2Seq} model. It takes as input a token sequence consisting of a latent variable value $\bm{h}_k$, a word sequence $\bm{x}$, and a relation symbol $r$, and predicts a word sequence representing the next event $\bm{z}$. We enrich the input vocabulary of the chosen \textsc{Seq2Seq} model with the symbols of $h_k$ and $r$ which are mapped to the corresponding latent embeddings and relation embeddings during forward propagation. 


We select ProphetNet~\cite{yan2020prophetnet} as the \textsc{Seq2Seq} backbone model, because it achieves the best performance on a number of natural language generation tasks. The encoder and the decoder of this model utilize $n$-stream self-attention mechanism and future $n$-gram prediction in order to encourage planning for the future tokens and prevent overfitting on strong local correlations. 
\paragraph{Training} The goal of training is to learn the parameters of the following model on ATOMIC,
\begin{align}
    \max_{\bm{\theta}_{\mathcal{B}}} \prod_{r(\bm{x}, \bm{z}) \in \mathcal{B}}\sum_{k=1}^K Pr(\bm{z}| h_k, \bm{x}, r; \bm{\theta}_{\mathcal{B}}) Pr(h_k | \bm{x}).
\end{align}
More specifically, we train the model on each \textit{if-then} relation of the form ``\textit{if} $\bm{x}$ \textit{and} $r$ \textit{then} $\bm{z}$'' in ATOMIC by taking $\bm{x}$ and $r$ as input and predicting $\bm{z}$.

Prior work on diverse machine translation~\cite{DBLP:conf/conll/HeHN18,shen2019mixture,DBLP:conf/emnlp/ChoSH19} suggests to apply online hard EM by interleaving the following two steps for each mini-batch.


\begin{itemize}
    \item \textbf{E-step}: estimate the value of the latent variable through $\hat{k} = \arg\max_{k} Pr(\bm{z}| h_k, \bm{x}, r; \bm{\theta}_{\mathcal{B}})$ using the current parameters $\bm{\theta}_{\mathcal{B}}$.

\item \textbf{M-step}: The model parameters $\bm{\theta}_{\mathcal{B}}$ are then updated by minimising the cross-entropy loss on $Pr(\bm{z}|h_{\hat{k}},\bm{x};\bm{\theta}_{\mathcal{B}})$.
\end{itemize}
However, the greedy search in the E-step may still assign the same latent variable value to different target sequences. To eliminate the problem, we modify and constrain the E-Step by requiring that, different target sequences of the same input need to be assigned  different latent variable values. More specifically, for each output set  $S_{r,\bm{x}} := \{j \ | \  \exists \ r(\bm{x}, \bm{z}_j) \in \mathcal{B}\}$  sharing the same input $\bm{x}$ and $r$ in a mini-batch, we tackle this problem by solving the following combinatorial optimization problem. 
\begin{align}\nonumber
& \max_{\{u_{j,k}\}} \sum_{j}\sum_{k} u_{j,k} \log Pr(\bm{z}_j|h_{k},\bm{x},{r};\bm{\theta}_{\mathcal{B}}) \\\nonumber
& s.t. \ \  \sum_k \sum_j u_{j,k} = |S_{r,\bm{x}}| \\\nonumber 
& \ \ \ \ \ \ \  \sum_{j \in S_{r,\bm{x}}}  u_{j,k} \leq 1, \hspace{4pt} \forall k
\end{align}
where $u_{j,k}$ is a binary variable indicating the alignment between $\bm{z}_j$ and $h_k$, and $|S_{r,\bm{x}}|$ is the number of target events in the output set. Here we set $K$ always larger than $|S_{r,\bm{x}}|$.

We solve the above problem by a heuristic-based search. We compute the log probability of $Pr(\bm{z}_j|h_{k},\bm{x},r;\bm{\theta}_{\mathcal{B}})$ for all combination of $k$ and $j$. Then we sort those log probabilities, and select $u_{j,k}$ satisfying the hard constraints in order. More details about  the algorithm can be found in the pseudo code in Algorithm \ref{alg:comb_opt}.

\begin{algorithm}[h]
\small
\SetKwInOut{Input}{input}\SetKwInOut{Output}{output}
\Input{ source $\bm{x}$, target $\{\bm{z}_j\}_{j=1}^J$, relation $r$, latent variable $\{h_k\}_{k=1}^K$ ($K >J$)} 
\Output{Model parameter $\theta_{\mathcal{B}}$}
empty list $\Gamma$\\
\While{ $j < M$}
{
        
        \While {$k < K$}
        {
            $ l_{j,k} := p(\bm{z}_j|h_k,\bm{x},r,;\theta_{\mathcal{B}})$; \\
            $\Gamma$ := \text{add}~($\bm{x}$,r,$\bm{z}_j$, $h_k$,$l_{j,k}$) \text{to} $\Gamma;$ \\
        }
} 

$\Gamma$ := Sort($\Gamma$) \text{ based on values of $l_{j,k}$} \\
create empty lists $\Delta$, $\zeta$, $\iota$ 

\Repeat{ \text{every }$\bm{z}_j$ \text{ is assigned with a }$h_k$}{
\For{$\bm{x}$,$r$,$\bm{z}_j$, $h_k$,$l_{j,k} \in \Gamma$}
{
\lIf{$\bm{z}_j \notin \zeta~\textbf{and}~h_k \notin \iota$}
{
$ \text{add}~(\bm{x}, r, \bm{z_j}, l_k)~\text{to}~\Delta$,  $\text{add}~\bm{z}_j~\text{to}~\zeta$, $\text{add}~h_k~\text{to}~\iota$
}
}
}

\For{$ \bm{x}, r, \bm{z_j}, l_k \in \Delta$}
{
Make forward and backward propagation w.r.t the training objective (Equation 5) \\
Update the model parameter ($\theta$)\\
}
\Return{$\theta$}

\caption{Conditional Mixture Model}
\label{alg:comb_opt}
\end{algorithm}








\subsection{Answer Scoring Module}

The main objective of the answer scoring module in Eq. \eqref{eq:model_score} is to assign a score to each answer candidate. For this purpose, Bosselut~\shortcite{bosselut2019zeroShotQA} proposed an averaged word probability approach. In this method, event prediction is considered as language modelling task, hence the score is defined as the average probability of generating each token of an event. However, this method is not theoretically grounded and largely relies on heuristics. Based on the probabilistic model in Eq.  \eqref{eq:model_score}, the true answers should be semantically similar to the last events in plausible reasoning paths derived from contexts and questions. Thus, the answer scoring module solely depend on the last events of reasoning paths.
\begin{align}
    Pr(\bm{a}|\bm{Z}) = Pr(\bm{a}|\bm{z}_T)
\end{align}
where $\bm{z}_T$ denotes the event generated at time step $T$. As the distribution is characterised by semantic similarity between the last events and answer candidates, we define the distribution as:
\begin{align}
    Pr(\bm{a} | \bm{z}_{T}) = \frac{\exp(-\gamma d(\bm{a}, \bm{z}_{T}))}{\sum_{\bm{a}' \in \mathcal{A}} \exp(-\gamma d(\bm{a}', \bm{z}_{T}))}
    \label{eq:scoring}
\end{align}
where $d(\bm{a},\bm{z})$ is a distance function between an answer and an event, and $\gamma$ is a hyperparameter adjusting the temperature.

After distilling the \textsc{Seq2Seq} model with information of ATOMIC, we plug the trained KB module and the answer module into the model. \model{CosMo} answers questions by applying Eq. \eqref{eq:model_score}. We apply beam search to find top-$k$ reasoning paths for each latent variable value up to a pre-specified number of hops $T$. The most plausible answer is the most probable reasoning path, whose last event achieves the highest similarity with the answer.

\section{Experiments}
\label{sec:experiment}
In this section, we report the evaluation of our model on zero-shot question answering. To this end, we use test set and development set of SocialIQA~\cite{sap2019socialiqa}. We evaluate the performance of our model with two variations. In zero-hop setup, we only consider generating clauses using \model{CosMo} based on the context of the question, and the answers are then scored against the generated clauses. In one-hop setting, we take a step further and generate more clauses using the generated clauses in the previous step. This approach helps us uncover more implicit context-dependent information. The final answer is chosen against the combination of all generated clauses. 
To further analyse the capability of our proposed model in terms of clause generation, we compare our model to other approaches on ATOMIC~\cite{sap2019atomic}, and test and development set of SocialIQA. 

\subsection{Datasets}
\paragraph{SocialIQA}~This dataset consists of commonsense questions, which aims to evaluate a model's capability in inferring implicit social context. Each question in this dataset is presented with a context, which describes the situation, and three answer candidates. For the purpose of addressing this task, we convert each question to one of the relations in the KG, using a pattern-based system. The details of this module is provided in the Appendix \ref{q2r}. This dataset contains a total of 37,588 questions. However, in a zero-shot setting we only use the development and test set for evaluation, where they contain 1,954 and 2,224 questions, respectively.

\paragraph{ATOMIC}~This dataset consists of 877K sets of subject, relation, and object, where each set describes a social commonsense situation. The subjects are an event (e.g., ``PersonX puts trust in PersonY"), which poses a social situation. The relations are categorised into 9 dimensions~(e.g., xEffect). The object is indicated by the relation, which shows the causes of subject, the effects of subject on the agent, and others, or attributes of the agent. The original data split, 710K/80K/87K for train/development/test, by Sap~\shortcite{sap2019atomic} is used in our experiments.

\subsection{Baselines}
For evaluating our proposed zero-shot question answering model, we compare \model{CosMo} to the state-of-the-art pretrained language models, GPT~\cite{radford2018improving}. Also, we consider different variation of GPT-2~\cite{radford2019language}, including GPT2-117M, GPT2-345M, and GPT2-762M. For this purpose the questions are converted to state sentence~(as described in Appendix B). The language model scores the answers based on cross-entropy loss of concatenation of context, question, and answers. In addition, we compare our model to two variation of COMET-CA and COMET-CGA~\cite{bosselut2019zeroShotQA}. We also report the performance of the state-of-the-art supervised methods, Bert~\cite{devlin2018bert} and RoBERTa~\cite{liu2019roberta}.

To further analyse the performance of our model in generation of clauses, we compare our model to the state-of-the-art automatic knowledge base completion model, that has been used in zero-shot commonsense question answering task, COMET~\cite{bosselut2019comet}. Also, to assess the diversity, we consider comparison of our proposed model to the \textsc{Seq2Seq} model presented in section \ref{sec:models}, without applying latent variable~(ProphetNet)\cite{yan2020prophetnet}, and state-of-the-art model for applying the latent variable~(MoE)\cite{shen2019mixture}. 

\subsection{Metrics}
For evaluating the performance of models in zero-shot question answering task, we report the accuracy of each model in choosing the correct answer. In addition, to compare the effectiveness of our proposed scoring function, we compare three variation of our scoring function to two different scoring functions proposed in \cite{bosselut2019zeroShotQA}. Furthermore, having a diverse clause generation is an advantage of our proposed model. As a quantitative evaluation, for a set of clause generated by the model denoted as $\{\hat{y}\}_{m=1}^M$, we use div\_bleu and div\_ngram~\cite{he2018sequence}, which are defined as follow:

\begin{align}
    \text{div\_bleu} &= 1 - (\sum_{i=1}^{M}\sum_{j=i+1}^{M}\text{BLEU}(\hat{y}_i,\hat{y}_j))/M(M-1)/2 \\
    \text{div\_ngram} &= 1 - \frac{|\cap_{m=1}^{M}\text{ngram}(\hat{y_m})|}{|\cup_{m=1}^{M}\text{ngram}(\hat{y_m})|}
\end{align}
where ngram(y) indicates the set of unique ngrams in a sequence $y$. For each model the top-50 clause generated by beam search is considered for evaluation purposes. For div\_bleu we report the average result of BLEU-1, BLEU-2, BLEU-3, and BLEU-4, with Smoothing1~\cite{chen2014systematic}. For div\_ngram, we report the average results of 1-gram, 2-gram, 3-gram, and 4-gram.

\subsection{Experimental Details}

For training our proposed Knowledge Graph Neuralisation model, \model{CosMo}~\footnote{Code available at https://github.com/farhadmfar/cosmo },  we finetune the \textsc{Seq2Seq} model using the pretrained model of Yan~\shortcite{yan2020prophetnet}. Our implementation is based on FAIRSEQ~\footnote{https://github.com/pytorch/fairseq}. The model consists of 12 layers of encoder and 12 layers of decoder. The embedding size and batch size are set to 1024 and 512, respectively. The number of future ngram is set to 2. We use Adam optimiser~\cite{kingma2014adam} with a peak learning rate of $1 \times 10^{-4}$. For the question answering module, we consider answering without taking a hop, and taking one-hop. Since we evaluate our model in zero-shot setting, the $\gamma$ in equation \ref{eq:scoring} is set to one, and we use cosine similarity function as the distance function. At each step, we consider beam-10 for clause generation.

\subsection{Results and Discussions}

\paragraph{Zero-shot Commonsense Question Answering} Table \ref{tab:qa} summarises the result of performance of different models in task of commonsense question answering. The performance of the models in zero-shot setting is reported in the top section of the table. It's clear that our proposed model outperforms the other methods, in both development and test set, by up to +5.2\%. Furthermore, the improvement over taking a hop in answering question, suggests that the implicit information related to a given context can help answering some questions. However, the existing gap with supervised models, indicates potential possibility of improvement. 

\begin{table}[h]
    \centering
    \begin{tabular}{l l c c}
         \tabularnewline\cline{2-4}\cline{2-4}\cline{2-4}
         & \textbf{MODEL} & \textbf{Dev Acc.} & \textbf{Test Acc.} 
         \tabularnewline\cline{2-4}\cline{2-4}\cline{2-4}
         \multirow{9}{*}{\small{\rotatebox[origin=c]{90}{Unsupervised}}} & Random & 33.3 & 33.3 \\
         & GPT~\cite{radford2018improving} & 41.8 & 41.7\\
         & GPT2-117M~\cite{radford2019language} & 40.7 & 41.5 \\
         & GPT2-345M~\cite{radford2019language} & 41.5 & 42.5 \\
         & GPT2-762M~\cite{radford2019language} & 42.5 & 42.4 
         \tabularnewline\cline{2-4}
         & COMET-CA~\cite{bosselut2019zeroShotQA} & 48.7 & 49.0 \\
         & COMET-CGA~\cite{bosselut2019zeroShotQA} & 49.6 & 51.9 
         \tabularnewline\cline{2-4}
         & \model{CosMo}~(zero-hop) & 52.9 & 53.1 \\
         & \model{CosMo}~(one-hop) & \textbf{54.8} & \textbf{55.0} 
         \tabularnewline\cline{2-4}\cline{2-4}\cline{2-4}
         \cline{2-4}\cline{2-4}\cline{2-4}
         \multirow{2}{*}{\small{\rotatebox[origin=c]{90}{Supervised}}}& BERT-Large~(supervised)~\cite{devlin2018bert} & 66.0 & 66.4 \\
         & RoBERTa~(supervised)~\cite{liu2019roberta} & 76.6 & 77.8 \tabularnewline\cline{2-4}\cline{2-4}\cline{2-4}
         & human & 86.9 & 84.4 
         \tabularnewline\cline{2-4}\cline{2-4}\cline{2-4}
    \end{tabular}
    \vspace{-1ex}
    \caption{The accuracy of answer prediction of our proposed model compared to the state-of-the-art models on SocialIQA, on development and test set.}
    \vspace{2ex}
    \label{tab:qa}
\end{table}

Table \ref{tab:sim-func} summarises the results of evaluation of different methods of scoring the answer candidates. The scoring function proposed by Bosselut~\shortcite{bosselut2019zeroShotQA}, averaged word probability, comes in two variations. To overcome the answer imbalance given specific questions, they propose adding Pointwise Mutual Information of question and answers to the original scoring function. The distance function~(Eq.  \eqref{eq:model_score}) in our proposed model is evaluated with three variations of directly using probability of \textsc{Seq2Seq} model~(\textsc{Seq2Seq}), BLEU function, and cosine distance. The results indicate that using cosine distance function improves the results by +4.93\% and +6.48\% over the second best approach, on development and test set, respectively. The experiments suggest that taking the similarity of clause generations at each step with the answers provides stronger classifier over answers, compared to considering only the probability of the generative models. The lack of performance of the latter configuration can be rooted in training phase of generative models, where some phrases, regardless of the context, are seen frequently together, resulting in achieving higher probability by the model. 

\begin{table}[h]
    \centering
    \begin{tabular}{l c c}
    \hline
    \textbf{MODEL} & \textbf{Dev Acc.} & \textbf{Test Acc.} \\ 
    \hline
    averaged word probability~\cite{bosselut2019zeroShotQA} & 36.59 & 33.67 \\
    averaged word probability~(without pmi)~\cite{bosselut2019zeroShotQA} & 34.98 & 35.05 \\
    \hline
    \model{CosMo} (\textsc{Seq2Seq}) & 34.98 & 35.05 \\
    \model{CosMo} (BLEU) & 47.93 & 46.62 \\
    \model{CosMo} (Cosine) & \textbf{52.86} & \textbf{53.1}\\
    \hline
    \end{tabular}
    \vspace{-1ex}
    \caption{The results of using our proposed answer classifier function compared to the baselines, on development and test set of SocialIQA.}
    \label{tab:sim-func}
\end{table}


\paragraph{Diverse Clause Generation}
One of the strength of our proposed model is the ability to generate diverse clauses given a subject and a relation. Table \ref{tab:div_ngram} shows the result of diverse generation in terms of div\_ngram and div\_bleu. As it can be seen, for div\_ngram, our proposed model achieves the highest performance on test set of ATOMIC. Furthermore, we observe that our model outperforms the baseline methods on development and test set of SocialIQA.

\begin{table}[h]
    \centering
    \small
    \begin{tabular}{c l c c c  }
    \cline{2-5} 
     & \textbf{MODEL} & \textbf{ATOMIC} & \textbf{SoicalIQA Dev.} & \textbf{SocialIQA Test}
    \tabularnewline\cline{2-5}
         \multirow{4}{*}{\small{\rotatebox[origin=c]{90}{div\_ngram}}} & COMET~\cite{bosselut2019comet}    & 43.19 & 41.50 & 39.91 \\
         & ProphetNet~\cite{yan2020prophetnet} & 67.27 & 49.36 & 49.39 \\
         & MoE~\cite{shen2019mixture} & 75.61 &  71.79 & 71.19\\
         & \model{CosMo} & \textbf{80.72} & \textbf{79.21} & \textbf{79.09}
    \tabularnewline\cline{2-5}
         \multirow{4}{*}{\small{\rotatebox[origin=c]{90}{div\_bleu}}} & COMET~\cite{bosselut2019comet}  & 84.22 & 84.03 & 79.44 \\
         & ProphetNet~\cite{yan2020prophetnet} & 82.44 & 77.58 & 74.55\\
         & MoE~\cite{shen2019mixture} & 89.60 & 88.33 & 87.90\\
         & \model{CosMo} & \textbf{93.03} & \textbf{92.34} & \textbf{92.39}
    \tabularnewline\cline{2-5}
    \end{tabular}
    \vspace{-1ex}
    \caption{The results of div\_ngram~(top section) and div\_bleu(bottom section) of our model compared to the baselines, on test set of ATOMIC and SocialIQA.}
    \label{tab:div_ngram}
\end{table}

The results of div\_bleu also shows that on ATOMIC and SocialIQA development and test set, our model outperforms all the baselines on all variation of BLEU function. The results suggest the capability of our proposed model in diverse clause generations.


\subsection{Qualitative Analysis}
In this section, we demonstrate the capability of our model on diverse clause generation, and its effect on commonsense question answering. Table \ref{tab:qual} provides two example from test set of SocialIQA. For each example, top-5 clauses generated by our proposed model and COMET, given context and question, are provided. Both examples show capability of our model in generating divers outputs, which results in finding the correct answer.

\begin{table}[h]
    \centering
    \small
    \begin{tabular}{|| l | p{10cm}||}
    \hline\hline
    \multirow{7}{*}{\small{\rotatebox[origin=c]{90}{Example 1}}} & \textit{Context}: Alex is a store owner and observed every person's contribution carefully. Alex rewarded every person accordingly. \\
       & \textit{Question}: Why did Alex do this? \\
       & \textit{Answers}: contribute to the local community, \\ 
       & close his store soon, \\ 
       & reward more for more deserving persons \\
       & \textit{Correct Answer}: reward more for more deserving persons\\
        \hline
        \multirow{5}{*}{\small{\rotatebox[origin=c]{90}{COMET}}} & to be a good employee\\
                         & to be helpful \\
                         & to be a good salesperson \\
                         & to make sure everything goes smoothly \\
                         & to be a good citizen \\
        \hline
        \multirow{5}{*}{\small{\rotatebox[origin=c]{90}{\model{CosMo}}}} &  to be fair \\
                             & to reward good work \\
                             & to appreciate good work \\
                             & to reward good people \\
                             & to show appreciation \\
        
    \hline\hline
         \multirow{2}{*}{\small{\rotatebox[origin=c]{90}{Example 2}}} & \textit{Context}: Bailey was a nice person so she called the family together. \\
           & \textit{Question}: What will happen to Others? \\
           & \textit{Answers}: talk to the family, hate bailey, thank bailey \\
           & \textit{Correct Answer}: thank bailey \\
          \hline
          \multirow{5}{*}{\small{\rotatebox[origin=c]{90}{COMET}}} & family members talk to Bailey \\
                        &    family members get to know Bailey \\
                        &     the family members get to know Bailey better \\
                        &     the family members talk to Bailey about Bailey \\
                        &     spend time with Bailey \\
        \hline
        \multirow{5}{*}{\small{\rotatebox[origin=c]{90}{\model{CosMo}}}} & the family members respect Bailey \\
                             & enjoy Bailey's company \\
                             & the family thank Bailey \\
                             & the family respects Bailey \\
                             & the people interact with Bailey \\
        \hline\hline
    \end{tabular}
    \caption{Examples of the test set of SocialIQA, with the clause generated by our proposed model, \model{CosMo}, compared to COMET.}
    \label{tab:qual}
\end{table}

\section{Conclusion}
\label{conclusion}
In this paper, we propose a novel approach for neuralising large-scale commonsense knowledge graph, Conditional \textsc{Seq2Seq}-based Mixture Model,~\model{CosMo}. Our proposed model provides diverse clause generation, to ensure coverage for the target task. We use the proposed model to generate diverse context related clauses, alongside with our proposed answer classifier model, to address the task of zero-shot commonsense question answering task. Empirical results on zero-shot commonsense question answering dataset show the superiority of our model over the state-of-the-art methods, by up to 5.2\%. Furthermore, our model outperforms baselines in terms of diversity of clause generations.

\clearpage
\bibliographystyle{coling}
\bibliography{reasoning.bib}

\clearpage
\section{Appendix}
\label{appendix}

\subsection{Appendix A}
\label{q2r}
\begin{table}[h]
\small
    \centering
    \begin{tabular}{l l c l}
    \hline
    \hline
         & Question &  Associated Relation & Assoicated phrase\\
        \hline\hline
        \multirow{2}{*}{cause} & Why did AGENT do this? & xIntent & AGENT did this because \dots \\
        & What does AGENT need to do before this? & xNeed & Before, AGENT needs to \dots \\
        \hline
        attribute & How would you describe AGENT? & xAttr & AGENT is \dots \\
        \hline
        \multirow{6}{*}{effect}& How would AGENT feel afterwards? & xReact & AGENT feels \dots \\
        & What will AGENT want to do next? & xWant & After, AGENT wants to \dots \\
        & What will happen to AGENT? & xEffect & The effect on AGENT will be \dots \\
        & How would others feel as a results? & oReact & Others feel \dots \\
        & What will others do next? & oWant & After, others will want to \dots \\
        & What will happen to others? & oEffect & The effect on other will be \dots \\
        \hline\hline
    \end{tabular}
    \caption{The templates that have been used to map the question from SocialIQA to the relations of ATOMIC, and phrases to use for language models, categorised by relation type~(cause/effect/attribute). ``AGENT" refers to the person who is the subject of the given context in the question answering tuple.}
    \label{tab:question2rel}
\end{table}

\end{document}